\documentclass[a4paper,twoside]{article}

\usepackage{epsfig}
\usepackage{subcaption}
\usepackage{calc}
\usepackage{amssymb}
\usepackage{amstext}
\usepackage{amsmath}
\usepackage{amsthm}
\usepackage{multicol}
\usepackage{pslatex}
\usepackage{apalike}
\usepackage{algorithm2e}
\usepackage[bottom]{footmisc}
\usepackage{graphicx}
\usepackage{enumitem}
\usepackage{tabularx}
\usepackage{float}
\usepackage{caption}
\usepackage{bm}
\usepackage{booktabs}     
\usepackage{array}        
\usepackage{xcolor}       
\usepackage{makecell}
\usepackage{pifont}
\newcommand{\cmark}{\ding{51}}
\usepackage{siunitx}
\usepackage{afterpage}
\usepackage{subcaption}
\usepackage{array}
\usepackage[breaklinks=true,bookmarks=false]{hyperref}
\usepackage{SCITEPRESS} 

\begin{document}
\newtheorem{prop}{Proposition}

\title{Self-Supervised Partial Cycle-Consistency for Multi-View Matching}

\author{\authorname{Fedor Taggenbrock\sup{1,2}\orcidAuthor{0009-0002-6166-0865}, Gertjan Burghouts\sup{1}\orcidAuthor{0000-0001-6265-7276} and Ronald Poppe \sup{2}\orcidAuthor{0000-0002-0843-7878}}
\affiliation{\sup{1}Utrecht University, Utrecht, Netherlands}
\affiliation{\sup{2}TNO, The Hague, Netherlands}
\email{\{fedor.taggenbrock, gertjan.burghouts\}@tno.nl, r.w.poppe@uu.nl}
}


\abstract{Matching objects across partially overlapping camera views is crucial in multi-camera systems and requires a view-invariant feature extraction network. Training such a network with cycle-consistency circumvents the need for labor-intensive labeling. In this paper, we extend the mathematical formulation of cycle-consistency to handle partial overlap. We then introduce a pseudo-mask which directs the training loss to take partial overlap into account. We additionally present several new cycle variants that complement each other and present a time-divergent scene sampling scheme that improves the data input for this self-supervised setting. Cross-camera matching experiments on the challenging DIVOTrack dataset show the merits of our approach. Compared to the self-supervised state-of-the-art, we achieve a 4.3 percentage point higher F1 score with our combined contributions. Our improvements are robust to reduced overlap in the training data, with substantial improvements in challenging scenes that need to make few matches between many people. Self-supervised feature networks trained with our method are effective at matching objects in a range of multi-camera settings, providing opportunities for complex tasks like large-scale multi-camera scene understanding.}

\onecolumn \maketitle \normalsize \setcounter{footnote}{0} 

\section{\uppercase{Introduction}}
Matching people and objects across cameras is essential for multi-camera understanding \cite{hao2023divotrack,multi_cam_und,HumanInteraction}. Matches are commonly obtained by solving a multi-view matching problem. One crucial factor that determines the quality of the matching is the feature extractors' generalization to varying appearances as a result of expressiveness and view angle~\cite{ristani2018features}. Feature extractors can be trained in a supervised setting, which requires labor-intensive data labeling \cite{hao2023divotrack}. The lack or scarcity of labeled data for novel domains is a limiting factor. Self-supervised techniques thus offer an attractive alternative because they can be trained directly on object and person bounding boxes without labels.

Effective, view-invariant feature networks have been learned with self-supervision through cycle-consistency, for use in multi-view matching, cross-view multi-object tracking, and re-identification (Re-ID) \cite{mvmhat,wang2020cycas}. Training these networks only requires sets of objects where there is a sufficient amount of overlap between sets of objects between views. For multi-person matching and tracking, sets are typically detections of people from multiple camera views \cite{mvmhat,hao2023divotrack}. When the overlapping field of view between cameras decreases in the training data, self-supervised cycle-consistency methods have a diluted learning signal.

\begin{figure*}[!htb]
\centering
\makebox[\textwidth]{\includegraphics[width=1\textwidth]{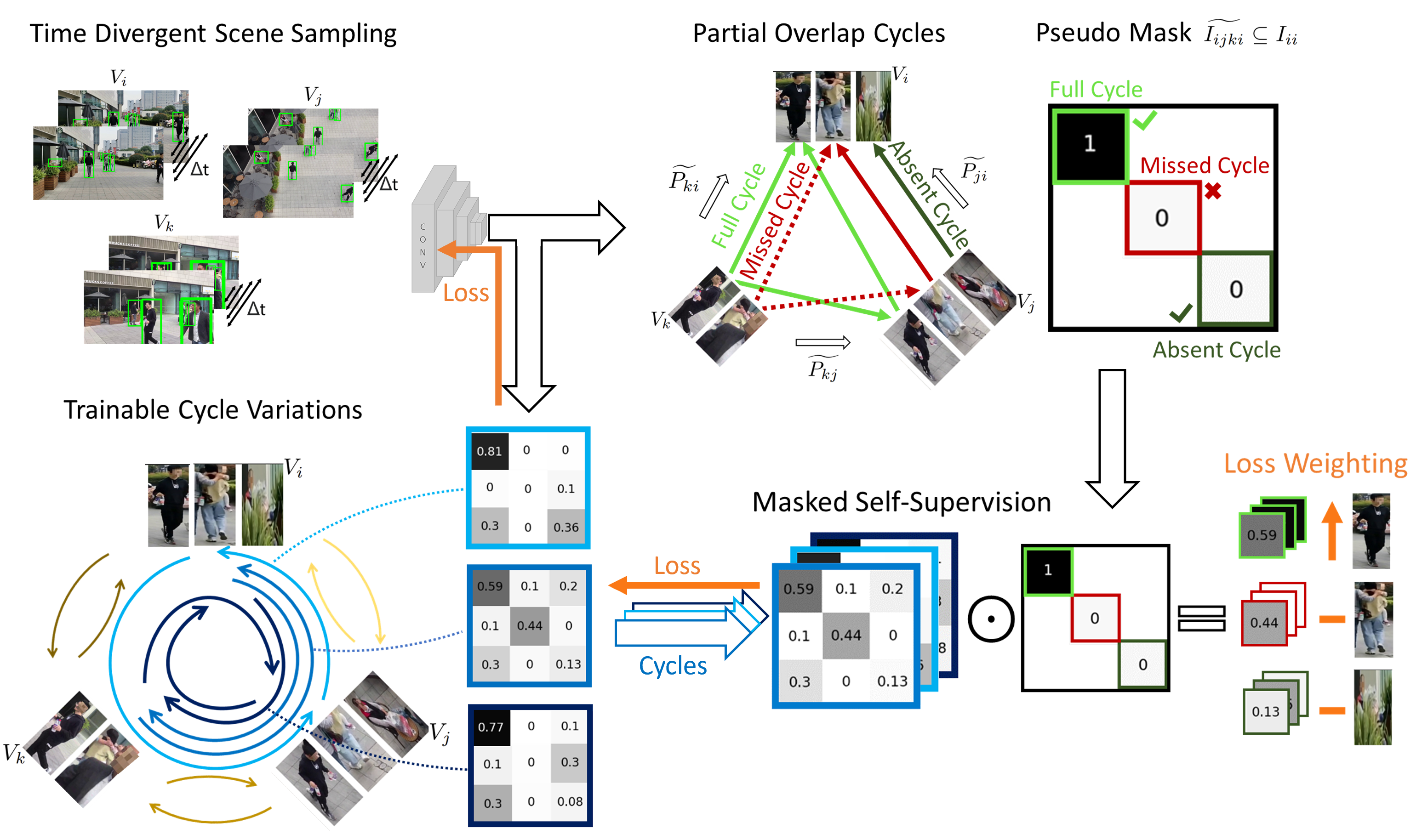}}
\captionsetup{justification=raggedright}
\caption{Overview of our self-supervised cycle-consistency training method. Trainable cycle variations (left bottom) are constructed from sampled batches (left top). Cycle matrices represent chains of matches starting and ending in the same view. With partial overlap, however, we construct a pseudo-mask of the identity matrix (top right) to determine which specific cycles should be trained due to partial overlap. This pseudo-mask is then used to provide a weighted loss signal with more emphasis on the positive predicted cycles (right bottom).} \label{fig:Method}
\end{figure*}

In this work, we address this situation and extend the theory of cycle-consistency for partial overlap with a new mathematical formulation. We then implement this theory to effectively handle partial overlap in the training data through a pseudo-mask, and introduce trainable cycle variations to obtain a richer learning signal, see Figure~\ref{fig:Method}. Consequently, we can get more out of the training data, thus providing a stronger cycle-consistency learning signal. Our method is shown to be robust in more challenging settings, with less overlap between cameras and fewer matches in the training data. It is especially effective for challenging scenes where few matches need to be found between many people. The additional information from partial cycle-consistency thus leads to substantial improvements, as shown in the experimental section. The code is also made open source \footnotemark
\footnotetext{For the open source code and theoretical analysis, see the Supplementary Materials available at \href{https://github.com/FedorTaggenbrock/Self-Supervised-Partial-Cycle-Consistency}{Github}.}



Our contributions are as follows:
\begin{enumerate}[label=\textbf{\arabic*.}, left=0pt, labelsep=1em, itemsep=0.5em]
    \item We extend the mathematical formulation of cycle-consistency to handle partial overlap, leading to a new formulation for partial cycle-consistency.
    \item We use pseudo-masks to implement partial cycle-consistency and introduce several cycle variants, motivating how these translate to a richer self-supervision learning signal. 
    \item We experiment with cross camera matching on the challenging DIVOTrack dataset, and obtain systematic improvements. Our experiments highlight the merits of using a range of cycle variants, and indicate that our approach is especially effective in more challenging scenarios.
\end{enumerate}

Section~\ref{sec: rel work} covers related works on self-supervised feature learning. Section~\ref{sec: theoretical extension} summarizes our mathematical formulation and derivation of cycle-consistency with partial overlap. Section~\ref{sec: SS with PCC} details our self-supervised method. We discuss the experimental validation in Section~\ref{sec: results and exp} and conclude in Section~\ref{sec: conclusion}.

\section{\uppercase{Related work}}\label{sec: rel work}
We first address the general multi-view matching problem, and highlight its application areas. Section~\ref{sec: sup match} summarizes supervised feature learning, whereas Section~\ref{sec: self sup match} details self-supervised alternatives. 

\subsection{Multi-View Matching}\label{sec: MVM}
Many problems in computer vision can be framed as a multi-view matching problem. Examples include keypoint matching~\cite{sarlin2020superglue}, video correspondence over time~\cite{palindrome_cyc}, shape matching~\cite{consistent_shapemaps_PSD}, 3D human pose estimation~\cite{multi_view_matching_fast_robust}, multi-object tracking (MOT)~\cite{sun2019dan}, re-identification (Re-ID)~\cite{AGW}, and cross-camera matching (CCM)~\cite{dan4ass}. Cross-view multi-object tracking (CVMOT) combines CCM with a tracking algorithm~\cite{mvmhat,hao2023divotrack}. The underlying problem is that there are more than two views of the same set of objects, and we want to find matches between the sets. For MOT, detections between two subsequent time frames are matched~\cite{deepsort}. Instead, in CCM, detections from different camera views should be matched. One particular challenge is that the observations have significantly different viewing angles. Such invariancies should be handled effectively through a feature extraction network. Such networks can be trained using identity label supervision but obtaining consistent labels across cameras is labor-intensive~\cite{hao2023divotrack}, highlighting the need for good self-supervised alternatives.  

\subsection{Supervised Feature Learning}\label{sec: sup match}
Supervised Re-ID methods~\cite{CT,AGW} work well for CCM. With labels, feature representations from the same instance are metrically moved closer, while pushing apart feature representations from different instances. Other approaches such as joint detection and Re-ID learning~\cite{hao2023divotrack}, or training specific matching networks~\cite{dan4ass} have been explored. Supervised methods for CCM typically degrade in performance when applied to unseen scenes, indicating issues with overfitting. Self-supervised cycle-consistency~\cite{mvmhat} has been shown to generalize better~\cite{hao2023divotrack}.

\subsection{Self-Supervised Feature Learning}\label{sec: self sup match}
Self-supervised feature learning algorithms do not exploit labels. Rather, common large-scale self-supervised contrastive learning techniques~\cite{simclr} rely on data augmentation. We argue that the significant variations in object appearance across views cannot be adequately modeled through data augmentations, meaning that such approaches cannot achieve view-invariance. Clustering-based self-supervised techniques~\cite{fan2018unsupervised} are also not designed to deal with significant view-invariance. Another alternative is to learn self-supervised features through forcing dissimilarity between tracklets within cameras while encouraging association with tracklets across cameras~\cite{li2019unsupervised}. Early work on self-supervised cycle-consistency has shown that this framework significantly outperforms clustering and tracklet based self-supervision methods ~\cite{wang2020cycas}. 
Self-supervision with cycle-consistency is especially suitable for multi-camera systems because it enables learning to associate consistently between the object representations from different cameras and at different timesteps. Trainable cycles can be constructed as series of matchings that start and end at the same object. Each object should be matched back to itself as long as the object is visible in all views. If an object is matched back to a different one, a cycle-inconsistency has been found which then serves as a learning signal~\cite{palindrome_cyc,wang2020cycas}. 

Given the feature representations of detections in two different views, a symmetric cycle between these two views can be constructed by combining two soft-maxed similarity matrices, matching back and forth. The feature network can then be trained by forcing this cycle to resemble the identity matrix with a loss~\cite{wang2020cycas}. This approach can be extended to transitive cycles between three views, which is sufficient to cover cycle-consistency between any number of views \cite{mvmhat,consistent_shapemaps_PSD}. With little partial overlap in the training data, forcing cycles to resemble the full identity matrix~\cite{mvmhat,wang2020cycas} provides a diluted learning signal that trains many non-existent cycles without putting proper emphasis on the actual cycles that should be trained. To effectively handle partial overlap, it is therefore important to differentiate between possibly existing and absent cycles in each batch. To this end, we implement a strategy that makes this differentiation. A work that was developed in parallel to ours~\cite{feng2024unveiling_new_mvmhat} has also found improvements with a related partial masking strategy. Our work confirms their observations that considering partial overlap improves matching performance. In addition, we provide a rigid mathematical underpinning for our method, introduce more cycle variations, and trace back improvements to characteristics of the scene including the amount of overlap between views.

Learning with cycle-consistency is not exclusive to CCM. Cycles between detections at different timesteps can be employed to train a self-supervised feature extractor for MOT~\cite{bastani2021self}, and cycles between image patches or video frames can serve to learn correspondence features at the image level \cite{dwibedi2019temporal,palindrome_cyc,time_cyc}. This highlights the importance of a rigid mathematical derivation of partial cycle-consistency in a self-supervised loss.

\begin{figure*}[!htb]
    \centering
    \hfill
    \begin{subfigure}{0.35\textwidth}
        \centering
        \includegraphics[width=\textwidth]{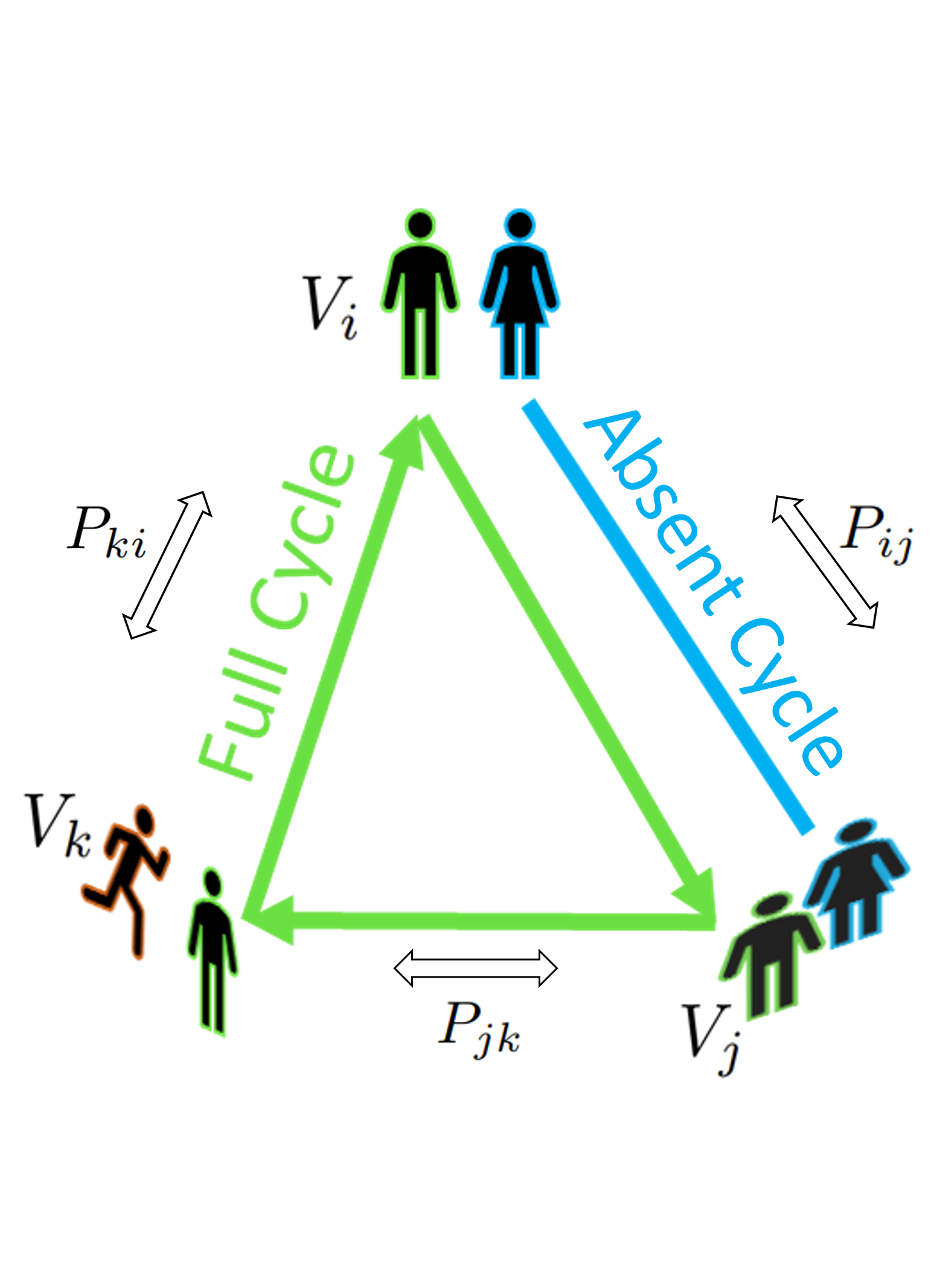} 
        \captionsetup{justification=raggedright}
    \end{subfigure}
    \hfill
    \begin{subfigure}{0.45\textwidth}
        \centering
        \includegraphics[width=\textwidth]{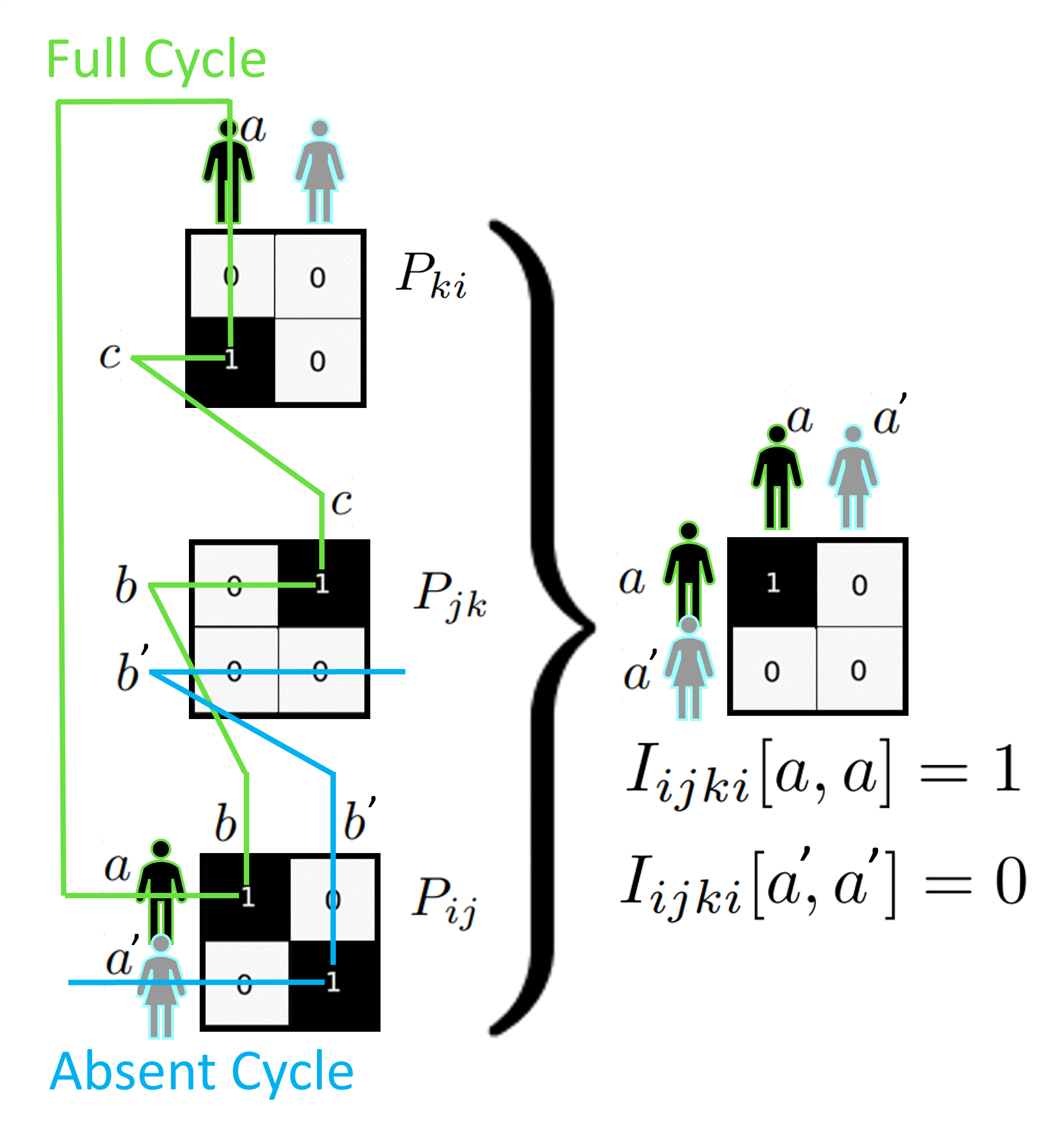}
        \captionsetup{justification=raggedright}
    \end{subfigure}
    \hfill
    \captionsetup{justification=raggedright}
    \vspace{-1em}
    \caption{Partial cycle-consistency and an interpretation of Equation~\ref{eq: def of I ijki summ}. $I_{ijki}[a,a]=1$ because $a$ is matched to $b$, matched to $c$ which is then matched back to $a$. The same does not hold for $a'$, so this cycle is absent.}
    \label{fig:overall_figure}
\end{figure*}

\newpage
\section{\uppercase{Partial Cycle-Consistency}} \label{sec: theoretical extension}
We summarize the main contributions from our theoretical extension of partial cycle-consistency, which appears in full in the supplementary materials\footnotemark[\value{footnote}]. Given are pairwise similarities $S_{ij}\in \mathbb{R}^{n_i\times n_j} \,\, \forall i,j$ between the views $V_i,V_j$, that contain $n_i,n_j$ bounding boxes. Partial multi-view matching aims to obtain the optimal partial matching matrices $P_{ij} \in \{0,1\}^{n_i\times n_j}\,\, \forall i,j,$ given the $S_{ij}$ that are partially cycle-consistent with each other. See also Figure~\ref{fig:overall_figure}. 
Partial cycle-consistency implies that, among others, matching from view $V_i$ to view $V_j$ and then to view $V_k$ should be a subset of the direct matching between $V_i$ and $V_k$. We make this subset relation explicit, pinpointing which matches get lost through view $V_j$ by inspecting the pairwise matches, proving equivalence to the original definition. We then prove that partial cycle-consistency in general implies the most usable form of self-supervision cycle-consistency, where matches are combined into full cycles that start and end in the same view and should thus be a subset of the identity matrix. We are able to explicitly define this usable form of partial cycle-consistency in proposition \ref{def:expl partial cyc cons summ}.
Based on this insight, in Section~\ref{sec: SS with PCC}, we construct subsets of the identity matrix during training to serve as pseudo-masks, improving the training process with partially overlapping views. Our explicit cycle-consistency proposition is as follows:

\newpage
\begin{prop}[Explicit partial cycle-consistency]\label{def:expl partial cyc cons summ}\text{}\\
    If a multi-view matching $\{ P_{ij} \}_{\forall i,j}$
    is partially cycle-consistent, it holds that:
    \begin{equation}\label{eq: explicit 1 summ}
    P_{ii} = I_{n_i \times n_i}  \quad \forall i \in \{1,\dots,N\},
    \end{equation}
    \begin{equation}\label{eq: explicit 2 summ }
    P_{ij}P_{ji} = I_{iji} \quad \forall i,j \in \{1,\dots,N\},
    \end{equation}
    \begin{equation}\label{eq: explicit 3 summ}
    P_{ij}P_{jk}P_{ki} = I_{ijki} \quad \forall i,j,k \in \{1,\dots,N\},
    \end{equation}

     where $ I_{iji} \subseteq I_{n_i \times n_i}$ is the identity map from view $i$ back to itself, filtering out matches that are not seen in view $V_j$: 
    \begin{equation}\label{eq: def of I iji summ}
        I_{iji}[a,c] =   \begin{cases} 
     1 & \text{if} \,\, a = c \,\,  \& \,\, \exists b \,\, \text{s.t.}\,\,  P_{ij}[a,b] = 1. \\
    0 & \text{else},
    \end{cases}
    \end{equation}
    
    and where $ I_{ijki} \subseteq I_{n_i \times n_i}$ is the identity mapping from view $i$ back to itself, filtering out all matches that are not seen in views $V_j$ and $V_k$: 
    \begin{equation}\label{eq: def of I ijki summ}
        I_{ijki}[a,d] =   \begin{cases} 
     1 & \text{if} \,\, a = d \,\, \& \,\, \exists b,c \,\, \text{s.t.}\,\,  P_{ij}[a,b]  \\
     & = P_{jk}[b,c] = P_{ki}[c,d] = 1. \\
    0 & \text{else}.
    \end{cases}
    \end{equation}
    The notation $X[\cdot, \cdot]$ is used for indexing a matrix $X$.
\end{prop}

The intuition behind Equation~\ref{eq: def of I ijki summ} can be best understood through the visualization in Figure~\ref{fig:overall_figure}. Here, $I_{ijki}[a', a'] = 0$ because there is a detection of $a'$ absent in view $V_k$, while $I_{ijki}[a, a] = 1$ because a full cycle is formed from the corresponding pairwise matches. The proofs with detailed explanations are given in the supplementary materials\footnotemark[\value{footnote}].

\newpage

\section{\uppercase{Self-Supervision with Partial Cycle-Consistency}}\label{sec: SS with PCC}
The theory of cycle-consistency and its relation to partial overlap can be translated into a self-supervised feature network training strategy. The main challenges are to determine which cycles to train, which loss to use, and how to implement the findings from Proposition~\ref{def:expl partial cyc cons summ} to handle partial overlap. Section~\ref{sec: Cons cyc} explores what cycles to train and how to construct them. Section \ref{sec: explicit partial loss} explores how to obtain partial overlap masks for the cycles that approximate the $I_{iji}, I_{ijki} \subseteq I_{n_i \times n_i}$ from Proposition~\ref{def:expl partial cyc cons summ}. It also explores how these masks can be incorporated in a loss to  deal with partial overlap during training.

\subsection{Trainable Cycle Variations}\label{sec: Cons cyc}
Given are the pairwise similarities $S_{ij}$ between all view pairs, obtained from the feature extractor $\phi$ that we wish to train. The idea is to combine softmax matchings of the $S_{ij}$ into cycles, similar to Equations~\ref{eq: explicit 2 summ } and \ref{eq: explicit 3 summ}. For this we use the temperature-adaptive row-wise softmax $f_{\tau}$ ~\cite{wang2020cycas} on a similarity matrix $S$ to perform a soft row-wise partial matching. This function has the differentiability needed to train a feature network and the flexibility to make non-matches for low similarity values. We get:
\begin{equation}
    f_{\tau}(S[a,b]) = \frac{\exp(\tau S[a,b])}{\sum_{b'} \exp(\tau S[a,b'])},
\end{equation}
where the notation $S[\cdot, \cdot]$ is used for matrix indexing. The temperature $\tau$ depends on the size of $S$ as in~\cite{wang2020cycas}. 

\subsubsection{Pairwise cycles.}
The pairwise cycles need to be constructed from just $S_{ij} = S_{ji}^T$. To this end, we take: 
\begin{equation}\label{eq: partial a_iji cycle}
    A_{ij} = f_{\tau}(S_{ij}), \quad A_{ji} = f_{\tau}(S_{ij}^T),  \quad  A_{iji} = A_{ij}A_{ji}.
\end{equation}
The pairwise cycle $A_{iji}$, originally proposed in \cite{wang2020cycas}, represents a trainable variant of the pairwise cycle $P_{ij}P_{ji}$ from Equation~\ref{eq: explicit 2 summ }, and so a learning signal is obtained by forcing it to resemble $I_{iji}$.  Note that the $A_{ij}$ and $A_{ji}$ differ because they match the rows and columns of $S_{ij}$, respectively. This is important because the loss then forces these different soft matchings to be consistent with each other, modelling the partial cycle-consistency constraint in Equation~\ref{eq: explicit 2 summ }. The loss will be the same for $A_{iji}$ and $A_{iji}^T = A_{ji}^TA_{ij}^T$, so just Equation~\ref{eq: partial a_iji cycle} suffices.

\subsubsection{Triplewise cycles.}
The triplewise cycles are constructed from $S_{ij}, S_{jk}$ and $S_{ki}$, and should resemble the $P_{ij}P_{jk}P_{ki}$ from Equation~\ref{eq: explicit 3 summ}. The authors in~\cite{feng2024unveiling_new_mvmhat} propose to use:
\begin{equation}\label{eq: cyc0 _first}
    A_{ijki}^0 = A_{ij}A_{jk}A_{ki},
\end{equation}
while in ~\cite{mvmhat}, the similarities are combined first so that:
\begin{equation}
    S_{ijk} = S_{ij}S_{jk},  \quad A_{ijk} = f_{\tau}(S_{ijk}),
\end{equation}
with which their triplewise cycle is created as:
\begin{equation}
    A_{ijki}^1 = A_{ijk}A_{kji}.
\end{equation}
We discovered that using multiple triplewise cycle constructions in the training improves the results. Each of the constructed cycles exposes a different inconsistency in the extracted features, so that a combination of cycles provides a robust training signal. We propose to use the following four triplewise cycles:
\begin{align}
    &A_{ijki}^0 = A_{ij}A_{jk}A_{ki}, \label{eq: cyc0} \\ 
    &A_{ijki}^1 = A_{ijk}A_{kji}, \label{eq: cyc1} \\
    &A_{ijki}^2 = A_{ijk}A_{ki}, \label{eq: cyc2} \\
    &A_{ijki}^3 = A_{ijk}A_{kij}A_{jki}. \label{eq: cyc3}
\end{align}
The cycles from Equation~\ref{eq: cyc1}-\ref{eq: cyc3} are also visualized in Figure~\ref{fig:Method} as the three blue swirls. In the following, $A_{ijki}$ can be used to refer to any of the four triplewise cycles in Equations~\ref{eq: cyc0}-~\ref{eq: cyc3}, and additionally $A_{iji}$ when assuming $j=k$. The symmetric property of the loss makes transposed versions of Equations~\ref{eq: cyc0} -~\ref{eq: cyc3} redundant. 


\subsection{Masked Partial Cycle-Consistency Loss}\label{sec: explicit partial loss}
The $A_{ijki}$ can be directly trained to resemble the identity matrix $I_{ii}$, by training each diagonal element in $A_{ijki}$ to be a margin $m$ greater than their corresponding maximum row and column values, similar to the triplet loss~\cite{wang2020cycas,mvmhat}. This is achieved through:
\begin{equation}\label{eq: L1}
    L_m(A_{ijki}) = \sum_{i=1}^{n_i}\text{relu}( \max_{\substack{b \neq a}}(A_{ijki}[a,b]) - A_{ijki}[a,a] +m ).
\end{equation}
The following loss enforces this margin over both the rows and columns:
\begin{equation}\label{eq: implicit loss}
    \mathcal{L}_{m}(A_{ijki}) = \frac{1}{2}(L_m(A_{ijki}) + L_m(A_{ijki}^T)).
\end{equation}
This loss, however, does not distinguish between absent and existing cycles that occur with partial overlap. Note that the ground truth $I_{ijki}$ are masks (or subsets) of the $I_{ii}$ that exactly filter out such absent cycles, while keeping the existing cycles, according to Equation~\ref{eq: def of I ijki summ} and visualized in Figure~\ref{fig:overall_figure}. In this figure, detections of the blue person form an absent cycle because the pairwise matches are not connected. The $I_{ijki}$ are constructed based on the ground truth matches $P_{ij}$. We therefore propose to construct pseudo-masks $\widetilde{I}_{ijki}$ from pseudo-matches $\widetilde{P}_{ij}$ that are available during self-supervised training. For this we use:
\begin{equation}
    \widetilde{P}_{ij} = \begin{cases}
    \left[ f_{\tau}(S_{ij}) \,\,\, > 0.5 \right] & \text{if} \,\, |V_i| < |V_j|, \\
    \left[f_{\tau}(S_{ij}^T)^T > 0.5\right] & \text{if} \,\, |V_j| < |V_i|, 
    \end{cases}
\end{equation}
where the Iverson bracket $\left[Predicate(X)\right]$ binarizes matrix $X$, with elements equal to 1 for which the predicate is true, and 0 otherwise. In $\widetilde{P}_{ij}$, each element in a view with fewer elements can be matched to at most one element in the other view, as desired for a partial matching. We construct the pseudo-masks as:
\begin{equation}\label{eq: pseudolabel definition}
    \widetilde{I}_{ijki}[a,a] =   \begin{cases} 
     1 & \text{if}  \,\,  \exists b,c \,\, \text{s.t.} \,\, \widetilde{P}_{ij}[a,b] \\
     &  =  \widetilde{P}_{jk}[b,c] = \widetilde{P}_{ki}[c,a] = 1. \\
    0 & \text{else}.
    \end{cases}
\end{equation}
$\widetilde{I}_{ijki}$ is invariant to the order in the $i,j,k$ sequence, and independent of the cycle variant for which it is used as a mask. Equation~\ref{eq: pseudolabel definition} can be vectorized as:

\begin{equation}
    \widetilde{I}_{ijki} =  \left[\widetilde{P}_{ij}\widetilde{P}_{jk}\widetilde{P}_{ki} \odot I_{ii} \geq 1\right].
\end{equation}

Our masked partial cycle-consistency loss extends the loss from Equation~\ref{eq: implicit loss} with the pseudo-masks $\widetilde{I}_{ijki}$, for which only the diagonal elements of predicted existing cycles are $1$. The absent cycles have diagonal elements of $0$. The loss uses two different margins $m_+ > m_{\emptyset} > 0$, where $m_+$ is used for cycles that are predicted to exist with $\widetilde{I}_{ijki}$, and $m_{\emptyset}$ is used for the cycles predicted to be absent:
\begin{equation}\label{eq: masked loss}
    \mathcal{L}_{explicit} = 
    \frac{\mathcal{L}_{m_+}( \widetilde{I}_{ijki}\odot A_{ijki}) + \mathcal{L}_{m_{\emptyset}}((I_{ii}-\widetilde{I}_{ijki})\odot A_{ijki})}{2}. 
\end{equation}

\newpage
\section{\uppercase{Results and Experiments}}\label{sec: results and exp}
We demonstrate the merits of a stronger self-supervised training signal from the addition of our cycle variations and partial cycle-consistency mask. We introduce the training setting, before detailing our quantitative and qualitative results.

\textbf{Dataset and metrics.}
DIVOTrack ~\cite{hao2023divotrack} is a large and varied dataset of time-aligned overlapping videos with consistently labeled people across cameras. 
The train and testset are disjoint sets with 9k frames from three overlapping camera's each. Three time-aligned overlapping frames are one matching instance. Frames from 10 different scenes are used, equally distributed over the train and test set. Our self-supervised feature network trains with the 9k matching instances of the trainset without its labels. We report the average cross-camera matching precision, recall and F1 score~\cite{dan4ass} over the 9k matching instances of the test set, averaged over five training runs with standard deviation. The average number of people per matching instance is around 19, but this varies per scene\footnotemark[\value{footnote}]. 

\textbf{Implementation details.}
Our contributions extends the state-of-the-art in self-supervised cycle-consistency ~\cite{mvmhat}. Our cycle variations from Equations~\ref{eq: cyc0}-\ref{eq: cyc3} are used instead of theirs, providing a diverse set of cycles to capture different cycle-inconsistencies. Previous methods without masking ~\cite{wang2020cycas}~\cite{mvmhat} use the loss from Equation~\ref{eq: implicit loss}. Our partial masking strategy instead constructs pseudo-masks with Equation~\ref{eq: pseudolabel definition} and uses these in our explicit partial masking loss from Equation~\ref{eq: masked loss}, with $m_+ =0.7$ and $m_{\emptyset} = 0.3$. We use the same training setup as~\cite{mvmhat} for fair comparison. Specifically, we use annotated bounding boxes without identity labels to extract features and train a ResNet-50~\cite{resnet} for 10 epochs with an Adam optimizer with learning rate $1e-5$. Matching inference uses the Hungarian algorithm between all view pairs, with an optimized partial overlap parameter to handle non-matches. 

\textbf{Time-divergent scene sampling.} 
Detections from multiple cameras at two timesteps are used in a batch such that cycles are constructed and trained between the pairs and triples for $2C$ views of the same scene, with $C$ the number of cameras~\cite{feng2024unveiling_new_mvmhat,mvmhat}. Time-divergent scene sampling gradually increases the interval $\Delta t$ between timesteps during training, with $\Delta t$ equal to the current epoch number. It also uses fractional sampling to obtain a balanced batch order, such that the local distribution of scenes resembles the average global distribution of scenes.

\begin{table*}[!htb]
    \centering
    \begin{tabular}{lc@{\hskip 5pt}c@{\hskip 10pt}cc@{\hskip 5pt}c@{\hskip -20pt}c@{\hskip -20pt}}
        \toprule
        \textbf{Model} & \multicolumn{3}{c}{\textbf{Standard}} & \multicolumn{3}{c}{\textbf{Time-Div. Scene Sampling}} \\
        \cmidrule(lr){2-4} \cmidrule(lr){5-7}
        & Precision & Recall & F1 & Precision & Recall & F1 \\
        \midrule
        MvMHAT~\cite{mvmhat} & 66.3 & 60.1 & 63.1$\pm$1.7 & 68.0 & 62.8 & 65.3$\pm$1.3\\
        Cycle variations (CV) & 68.8 & \scalebox{0.95}{\textbf{61.1}} & 64.7$\pm$1.9  & 70.4 & 62.3 & 66.1$\pm$1.4\\
        CV + Partial masking & \scalebox{0.95}{\textbf{71.0}} & 61.0 & \scalebox{0.95}{\textbf{65.6}}$\pm$\scalebox{0.95}{\textbf{1.1}} & \scalebox{0.95}{\textbf{71.7}} & \scalebox{0.95}{\textbf{63.6}} & \scalebox{0.95}{\textbf{67.4}}$\pm$\scalebox{0.95}{\textbf{0.9}} \\
        \bottomrule
    \end{tabular}
    \captionsetup{justification=raggedright}
    \caption{Cycle variations and partial masking together improve the overall matching performance by 2.5-2.1 percentage points. Every method benefits from time-divergent scene sampling, and combining everything boosts the previous SOTA by 4.3 percentage points, also improving stability.}
    \label{tab:ablation_study}
\end{table*}

\newpage
\subsection{Main Results}
We show the overall effectiveness of our cycle variations and partial masking as additions to the existing SOTA within the framework of self-supervised cycle-consistency ~\cite{mvmhat} in Table~\ref{tab:ablation_study}.

The first paper in this framework \cite{wang2020cycas} used a simple baseline approach with just pairwise cycles, and showed the effectiveness compared to multiple other self-supervised feature learning methods using clustering ~\cite{fan2018unsupervised} and tracklet based techniques ~\cite{li2019unsupervised} among others. The authors in ~\cite{mvmhat} and \cite{feng2024unveiling_new_mvmhat} expanded upon this framework, where \cite{feng2024unveiling_new_mvmhat} is not open source. We report the results in our paper both with and without time-divergent scene sampling, as this simply makes the data input richer, improving performance regardless of which cycle-consistency method is used. We find that combining cycle variations, partial masking and Time Divergent Scene Sampling boosts the F1 matchings score of the previous SOTA by 4.3 percentage points, and that this combination is also the most consistent of all approaches. 
To put the results of Table \ref{tab:ablation_study} in perspective, we report that the F1 matching score of a Resnet pretrained on ImageNet is $16.8$, while a supervised  SOTA  Re-ID model \cite{AGW} with an optimized network architecture and hard negative mining is able to obtain a matching score of $82.28$. This illustrates the strength of self-supervised cycle-consistency in general, showcasing its ability to significantly improve the feature quality of a pretrained ResNet. It also shows that our unoptimized self-supervised method is not to far from an optimized supervised baseline.

\subsubsection{Results per scene.}
The 10 scenes in the train and test data provide different challenges. During training, scenes with little overlap provide a worse learning signal for the overall model. During testing, scenes that require few matchings between many people are significantly more challenging. Insights into the overlap and number of people per scene is provided in the supplementary materials\footnotemark[\value{footnote}]. The scenes Ground, Side and Shop contain the highest number of people, around 24-32 per frame on average. The scenes Side and Shop also have little overlap, so that few matches needed to be correctly found from many possible ones. These scenes can thus be considered as the most challenging test set scenes. Table~\ref{tab:per_scene results} reports the matching results per scene. Our methods outperform \cite{mvmhat} on every test set, with the largest (relative) gains on Ground, Side and Shop, with 9.1, 5.6 and 4.7 percentage points, respectively, highlighting the improved expressiveness of our feature network.

\begin{table*}[!htb]
    \centering
    \begin{tabular}{lccccc}
    \toprule
    \textbf{Methods} &  \textbf{Gate2} & \textbf{Square} & \textbf{Moving} & \textbf{Circle} & \textbf{Gate1} \\
    \midrule
    MvMHAT~\cite{mvmhat} & 88.1 & 73.3 & 73.1 &67.4 & 67.2\\
    Ours w\textbackslash o Masking & \textbf{88.3(+0.2)}  & \textbf{74.9(+1.6)} & 74.9(+1.8) & 68.7(+1.3) & 69.6(+2.4) \\
    Ours & \textbf{88.3(+0.2)}  & \textbf{74.9(+1.6)} & \textbf{76.2(+3.1)}  & \textbf{69.9(+2.5)} & \textbf{70.4(+3.2)} \\
    \midrule 
    \textbf{Methods} & \textbf{Floor} & \textbf{Park} & \textbf{Ground} & \textbf{Side} & \textbf{Shop} \\
    \midrule
    MvMHAT~\cite{mvmhat} &64.7 & 58.2 & 56.9 & 56.0 & 42.1 \\
    Ours w\textbackslash o Masking & 65.2(+0.5) & 58.4(+0.2) &64.5(+7.6) & 58.9(+2.9) & 45.5(+3.4)\\
    Ours & \textbf{66.8(+2.1)} & \textbf{60.4(+2.2)} &\textbf{66.0(+9.1)} & \textbf{61.6(+5.6)} & \textbf{46.8(+4.7)}\\
    \bottomrule
    \end{tabular}
    \captionsetup{justification=raggedright}
    \caption{Results per scene. Our methods improve the average F1 score on every scene. Crowded challenging test scenes like \textbf{Ground}, \textbf{Side} and \textbf{Shop} benefit most, with improvements of 9.1, 5.6 and 4.7 percentage points respectively.}
    \label{tab:per_scene results}
        \vspace{3em}
\end{table*}

\subsubsection{Partial overlap experiments.} We experiment with artificially reducing the field of view in the training data by 20-40\%. We implement this by reducing the actual width of each camera view starting from the right, throwing away the bounding boxes outside this reduced field of view. We train on these reduced overlap datasets and measure the robustness for each method, because self-supervision through cycle-consistency learns from overlap. An overlap analysis for the original and reduced datasets is provided in Table~\ref{tab: overlap analysis}, and the evaluation results when training with the reduced data are shown in Table~\ref{tab: overlap results}. We observe that our method is robust and contributes to the performance even in these harder training scenarios.

\begin{table*}[!htb]
    \centering
    \begin{tabular}{lc@{\hskip 10pt}c@{\hskip 10pt}c}
        \toprule
        \textbf{Jaccard Index} & \textbf{Full Train\textbar Test} & \textbf{80\% Train Overlap } & \textbf{60\% Train Overlap }   \\
        \midrule
        \textbf{Two Cameras} & 0.40\textbar0.38 & 0.37 & 0.29 \\
        \textbf{Three Cameras} & 0.26\textbar0.23 & 0.24 &  0.15   \\
        \midrule
        \textbf{Num People} & 18.4\textbar19.4 & 16.5 & 14.0 \\
        \bottomrule
    \end{tabular}
    \captionsetup{justification=raggedright}
    \caption{The original train dataset has an average of 40\% IoU between any two cameras, 26\% people visible in all three cameras, and 18.4 unique people per frame. We reduce the FOV to simulate  harder train data with less overlap.}
    \label{tab: overlap analysis}
    \vspace{3em}
\end{table*}

\begin{table*}[!htb]
    \centering
    \begin{tabular}{lc@{\hskip 15pt}c@{\hskip 10pt}c}
        \toprule
         & \textbf{Full Train} & \textbf{80\% Train Overlap} & \textbf{60\% Train Overlap} \\
        \midrule 
        \textbf{Methods} & \multicolumn{3}{c}{\hspace{-2em}\textbf{test set F1 score}} \\
        \midrule
        MvMHAT~\cite{mvmhat} & 63.1$\pm$1.7 & 60.6$\pm$1.6 & 55.0$\pm$2.3    \\
        Ours w\textbackslash o Masking & 66.1$\pm$1.4 & 63.0$\pm$1.9 &   56.5$\pm$2.3  \\
        Ours &  $\bm{67.4}$$\pm$$\bm{0.9}$  & $\bm{63.8}\pm\bm{1.2}$ & $\bm{57.9}\pm\bm{1.5}$  \\
        \bottomrule
    \end{tabular}
    \captionsetup{justification=raggedright}
    \caption{Our methods consistently improve performance, even with sparser training data that is reduced in partial overlap.} 
    \label{tab: overlap results}
    \vspace{3em}
\end{table*}

\subsubsection{Cycle variations ablation.}
Our cycle variations use Equations~\ref{eq: cyc0}-~\ref{eq: cyc3} to construct multiple trainable cycles to obtain a richer learning signal. We perform an ablation study on the effectiveness of each cycle, with and without masking, in Table~\ref{tab: cyc abl.}. We find that our new $A_{ijk}A_{ki}$ and $A_{ijk}A_{kij}A_{jki}$ cycles from Equations~\ref{eq: cyc2} and ~\ref{eq: cyc3} perform well on their own and even better when combined with the cycles from Equations ~\ref{eq: cyc0} and ~\ref{eq: cyc1}. We observe that multiple cycle variations work especially well in the presence of masking, showing that these methods partly complement each other.


\begin{table*}[!htb]
    \centering
    \label{tab: cycle ablation}
    \begin{tabular}{c@{\hskip 10pt}c@{\hskip 10pt}c@{\hskip 10pt}c@{\hskip 10pt}c@{\hskip 10pt}c}
        \includegraphics[width=0.092\textwidth]{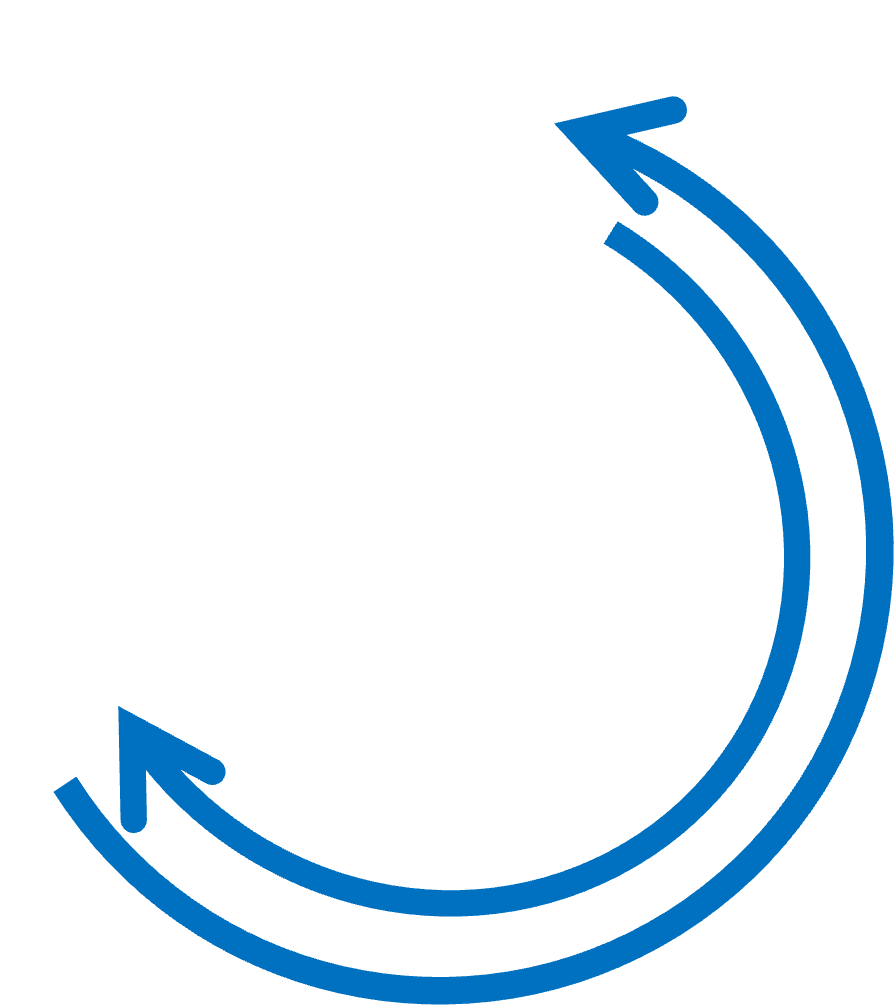} & \includegraphics[width=0.098\textwidth]{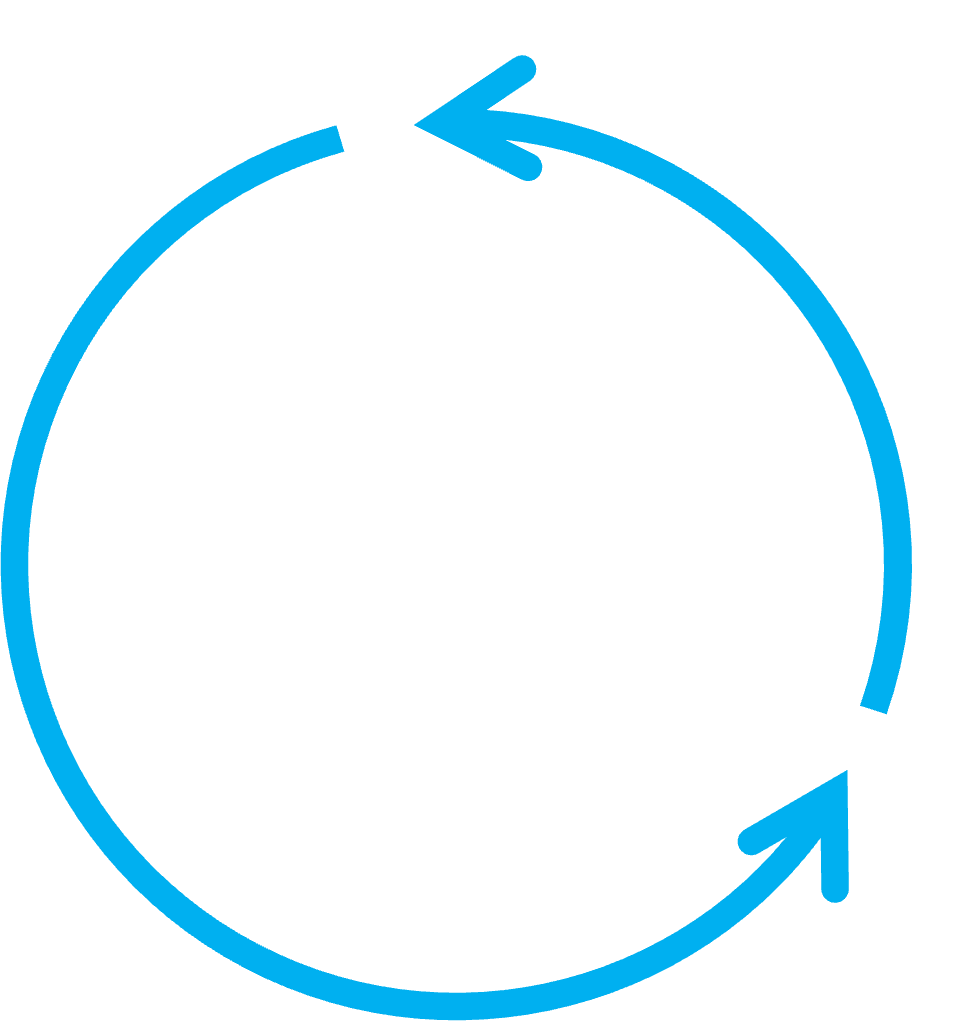} & \includegraphics[width=0.1\textwidth]{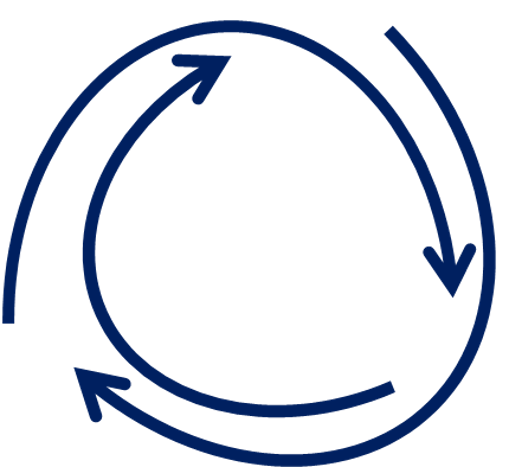} & \includegraphics[width=0.1\textwidth]{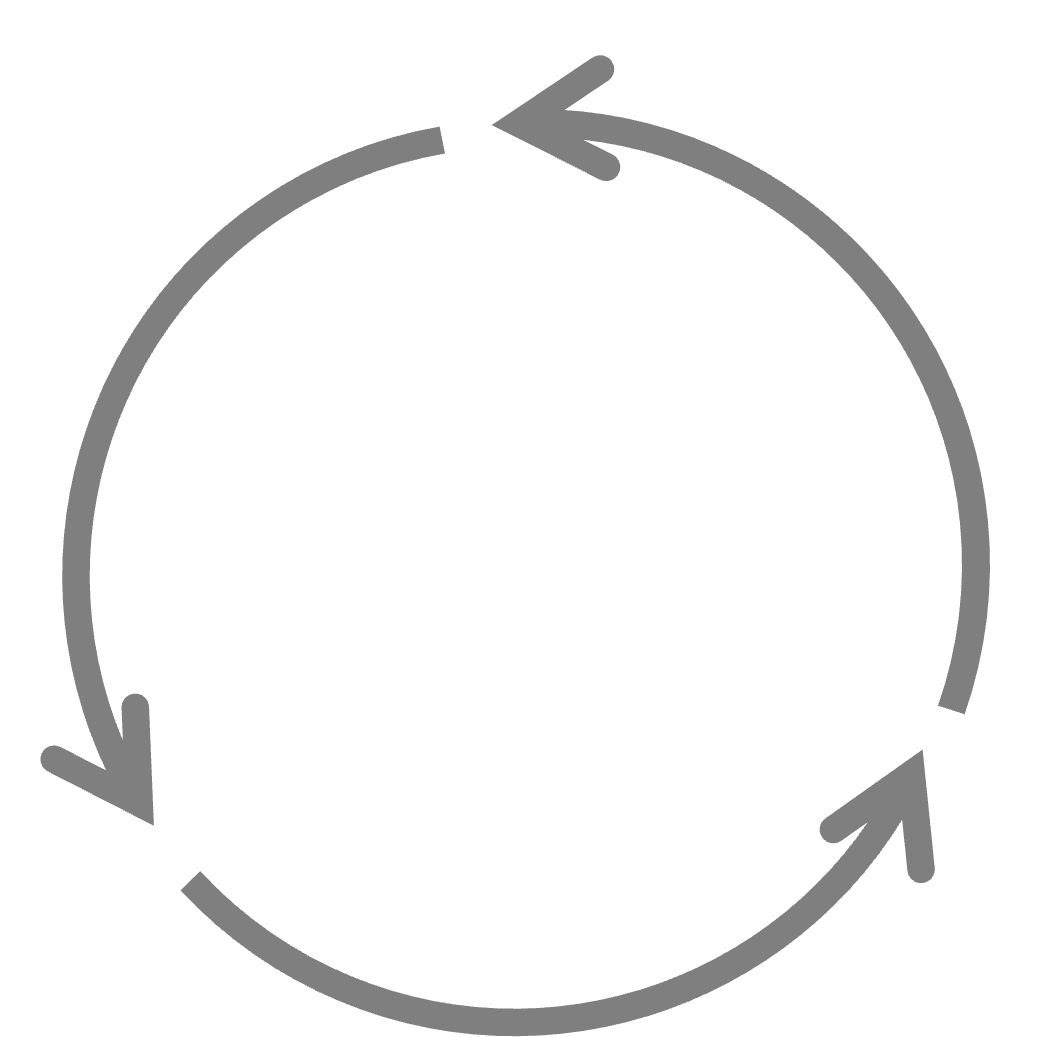}  \\
        \toprule
          $A_{ijk}A_{kji}$ & $A_{ijk}A_{ki}$   & $A_{ijk}A_{kij}A_{jki}$   & $A_{ij}A_{jk}A_{ki}$  & \textbf{w\textbackslash o} & \textbf{with} \\
           Eq~\ref{eq: cyc1} \cite{mvmhat} & Eq~\ref{eq: cyc2} [Ours] & Eq~\ref{eq: cyc3} [Ours] &  Eq~\ref{eq: cyc0} \cite{feng2024unveiling_new_mvmhat}&  \textbf{Masking} & \textbf{Masking} \\
        \midrule
           \cmark & & & & 65.1 $\pm$ 0.9 & 66.7$\pm$ 0.6 \\ 
           & \cmark & & & $\bm{66.4} \pm \bm{1.0}$ & 66.2$\pm$1.5\\ 
           & & \cmark  & & 65.6 $\pm$ 1.8& 66.4$\pm$1.2\\ 
           & & & \cmark & 57.7 $\pm$ 1.5& 55.9$\pm$1.2 \\ 
        \midrule
          \cmark & \cmark & & & 65.6 $\pm$ 1.5 & 66.7 $\pm$ 0.9 \\ 
          \cmark & \cmark & \cmark & &66.2 $\pm$  1.2& 66.9 $\pm$ 0.7 \\ 
          \cmark & \cmark & \cmark & \cmark & 66.3 $\pm$ 1.0 &  $\bm{67.2}\pm\bm{1.1}$\\ 
        \bottomrule
    \end{tabular}
    \captionsetup{justification=raggedright}
    \caption{Ablation of the cycle variations, also linking the Equations with illustrations. Our new $A_{ijk}A_{ki}$ and $A_{ijk}A_{kij}A_{jki}$ cycles work well individually, and even better in combination with the other cycle variations. Partial masking is also most effective when combined with multiple cycle variations. Our final method uses the setup from the bottom row with masking.}
    \label{tab: cyc abl.}
\end{table*}

\begin{figure*}[!htb]
\vspace{2em}
\centering
\includegraphics[width=0.9\textwidth]{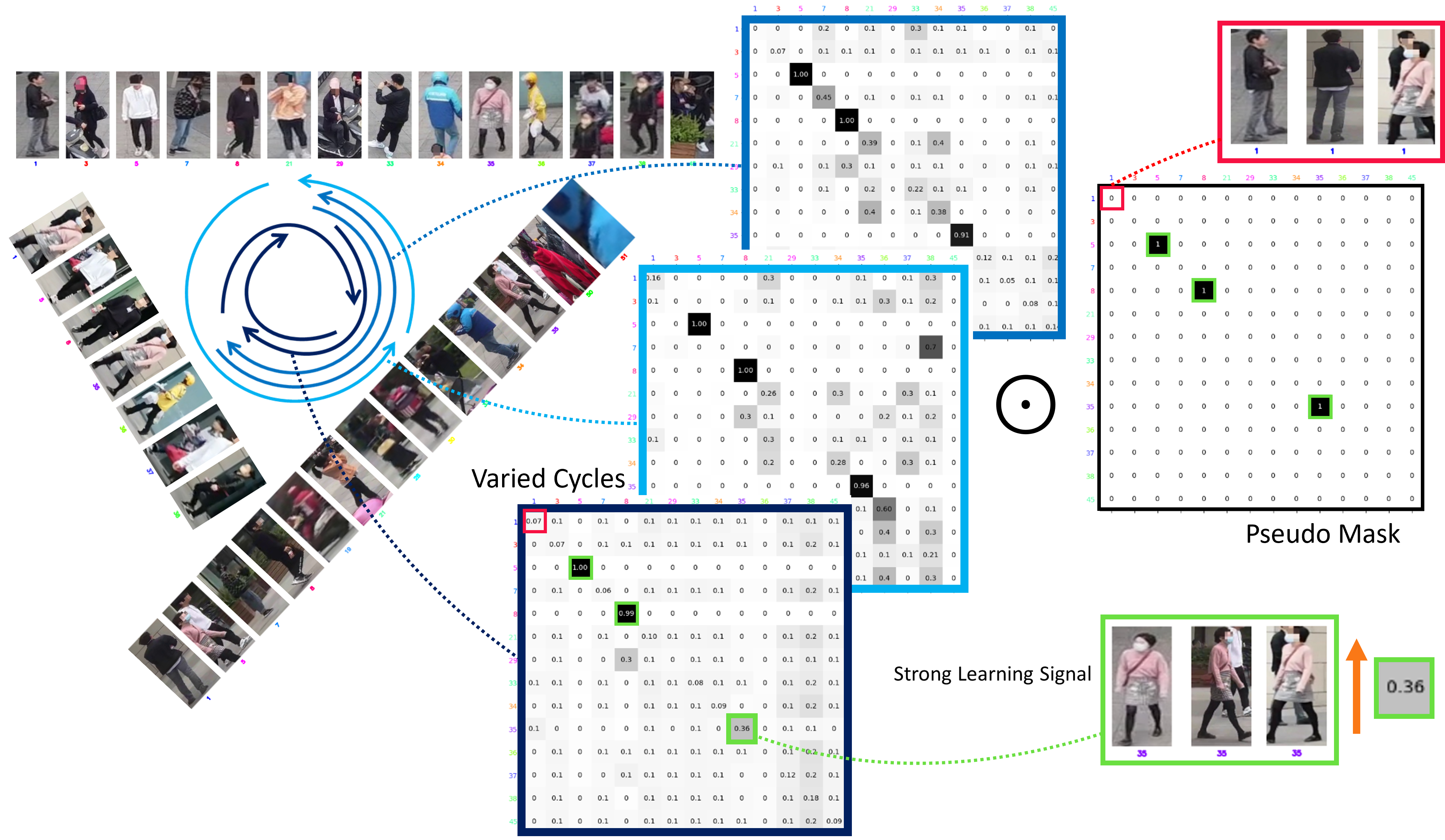}
\captionsetup{justification=raggedright}
\caption{Qualitative example during training. Each of the blue swirls, representing Equations~\ref{eq: cyc1}-\ref{eq: cyc3}, constructs a cycle matrix with various cycle-inconsistencies. Partial overlap requires that only some of the diagonal elements are trained as cycles. The pseudo-mask correctly finds the existing cycles, except for a heavily occluded one. A strong learning signal is obtained from one of the diagonals of the dark blue cycle.}
 \label{fig:train inzicht}
\end{figure*}

\begin{figure*}[!htb]
    \centering
    \begin{subfigure}[b]{0.85\textwidth}
        \centering
        \includegraphics[width=\textwidth]{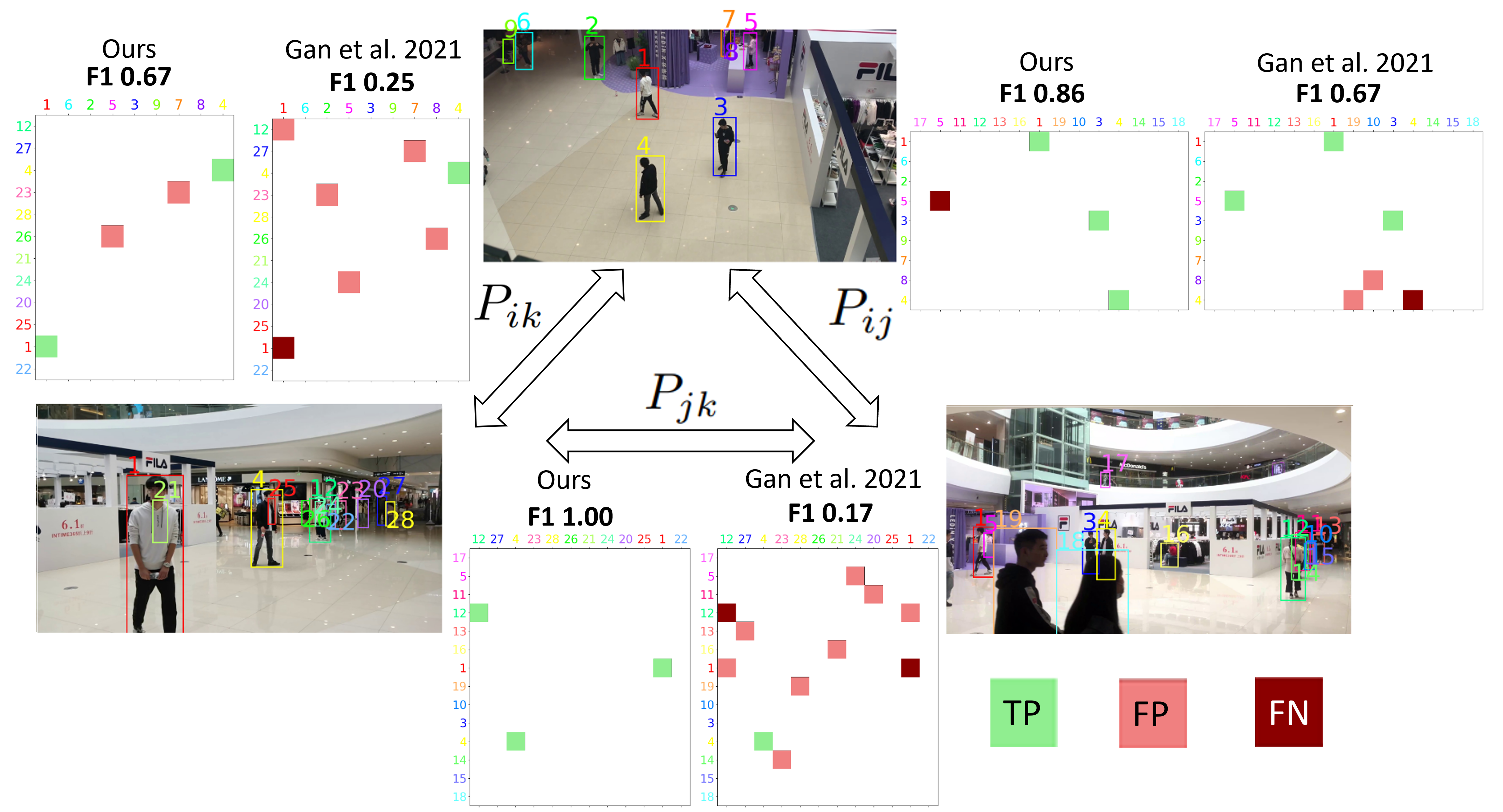}

        \label{fig: test res A}
    \end{subfigure}
    
    \begin{subfigure}[b]{0.85\textwidth}
        \centering
        \includegraphics[width=\textwidth]{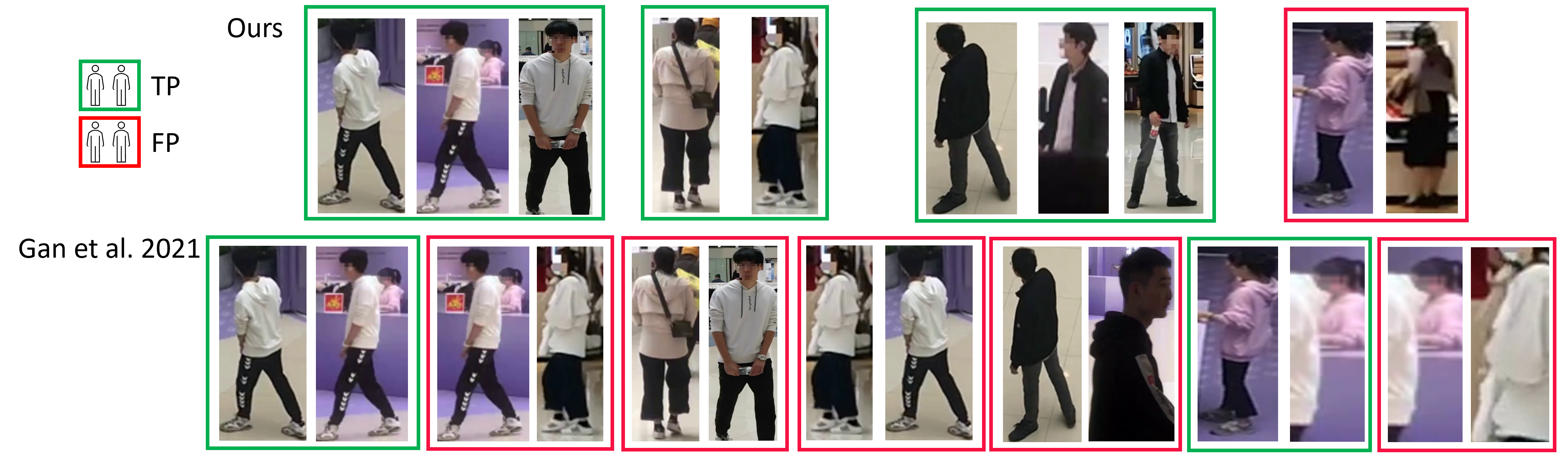}
        \label{fig: test res B}
    \end{subfigure}
    \captionsetup{justification=raggedright}
    \caption{Qualitative example during matching inference for a difficult frame in the test set. Our model is able to match with significantly fewer false positives. The matches found with our method are based on subtle clothing details, and have been correctly found in the presence of significant view angle differences and occlusion, significantly improving over the previous SOTA}
    \label{fig: overall test res}
    \vspace{1em}
\end{figure*}

\subsection{Qualitative Results}
Figure~\ref{fig:train inzicht} illustrates the contribution of the various cycles and pseudo-mask during training. In this specific example, it can be seen how the varied cycle constructions are cycle-inconsistent in different ways. Consequently, a robust learning signal is obtained from combining these cycle variants. The figure also shows the pseudo-mask $I_{ijki}$ that is constructed for this batch, where the existing cycles are correctly found with the exception of a severely occluded one in the top left, which you would not want to train anyway. The low value of 0.36 on the diagonal of the dark blue cycle means that a strong self-supervised learning signal is obtained from the masking, forcing the model to output more similar features for the different views of the person in pink. 

Figure~\ref{fig: overall test res} provides insight into the test set matching performance of our model compared to \cite{mvmhat}. It shows how our model effectively finds the pairwise matches at test time in a crowded scene. Note the difficulty of the matching problem, and how our method has significantly fewer false positive matches. The figure also demonstrates that our method is able to match significantly different representations of the same person across cameras, while differentiating between very similar looking people based on subtle clothing details.

\newpage
\newpage

\section{\uppercase{Conclusions}} \label{sec: conclusion}
We have extended the mathematical formulation of cycle-consistency to partial overlaps between views. We have leveraged these insights to develop a self-supervised training setting that employs multiple new cycle variants and a pseudo-masking approach to steer the loss function. The cycle variants expose different cycle-inconsistencies, ensuring that the self-supervised learning signal is more diverse and therefore stronger. We also presented a time divergent batch sampling approach for self-supervised cycle-consistency. Our methods combined improve the cross-camera matching performance of the current self-supervised state-of-the-art on the challenging DIVOTrack benchmark by 4.3 percentage points overall, and by 4.7-9.1 percentage points for the most challenging scenes.

Our method is effective in other multi-camera downstream tasks such as Re-ID and cross-view multi-object tracking. One limitation of self-supervision with cycle-consistency is its dependence on bounding boxes in the training data. Detections from an untrained detector could be used to train with instead, but this would likely degrade performance. Another area for improvement is to take location and relative distances into account both during training and testing, as this provides informative identity information.

Self-supervision through cycle-consistency is applicable to many more settings than just learning view-invariant object features. We believe the techniques introduced in this paper also benefit works that use cycle-consistency to learn image, patch, or keypoint features from videos or overlapping views.

\bibliographystyle{apalike}
{\small
\bibliography{egbib}}

\end{document}